\documentclass[letterpaper, 10 pt, conference]{ieeeconf}

\IEEEoverridecommandlockouts
\usepackage{cite}
\usepackage{booktabs}
\usepackage{amsmath,amssymb,amsfonts}
\usepackage{graphicx}
\usepackage{array}
\usepackage[ruled,vlined,linesnumbered]{algorithm2e}
\usepackage{textcomp}
\usepackage{url}
\title{\LARGE \bf
Tether-Aware Dynamic Collision Avoidance for USV-HROV Systems
}

\author{Yang Gu, Ziyang Hong, Xuanlin Chen, Hao Wei, Cheng Wang, Shujie Yang and Yulin Si%
\thanks{This work has been submitted to the IEEE for possible publication. Copyright may be transferred without notice, after which this version may no longer be accessible. Corresponding author: Yulin Si (yulinsi@zju.edu.cn).}%
}

\begin{document}

\maketitle
\thispagestyle{empty}
\pagestyle{empty}


\begin{abstract}
Heterogeneous marine robotic systems composed of an unmanned surface vehicle (USV) and a hybrid remotely operated vehicle (HROV) have shown great potential for subsea cable inspection. 
In such missions, the USV tracks the HROV at the surface while supplying power and communication through an umbilical tether. 
However, dynamic collision avoidance for the USV during HROV tracking is challenging because the submerged tether may scrape against passing vessels, while evasive maneuvers can enlarge the USV--HROV separation, thereby increasing the likelihood of tether tautness and compromising HROV operations.
To address these challenges, this work proposes a tether-aware dynamic collision avoidance method for a USV tracking an HROV. 
First, a tether safety-aware planar domain is introduced to represent the three-dimensional collision risk between the tether and obstacle vessels without an explicit tether shape model. 
Second, a tether tautness-aware velocity obstacle method is developed to achieve safe avoidance while reducing the likelihood of tether tautness. 
Finally, the method is integrated with line-of-sight guidance to coordinate HROV tracking and collision avoidance. 
Gazebo-based simulations show that the proposed method avoids dynamic obstacle vessels while maintaining tether safety and reducing the likelihood of tether tautness during USV evasive maneuvers.
Code will be released at \url{https://github.com/For-RAL-Manuscript/Tether-Aware-Dynamic-Collision-Avoidance.git}.
\end{abstract}


\section{Introduction}
\begin{figure}[!t]
	\centerline{\includegraphics[width=\columnwidth]{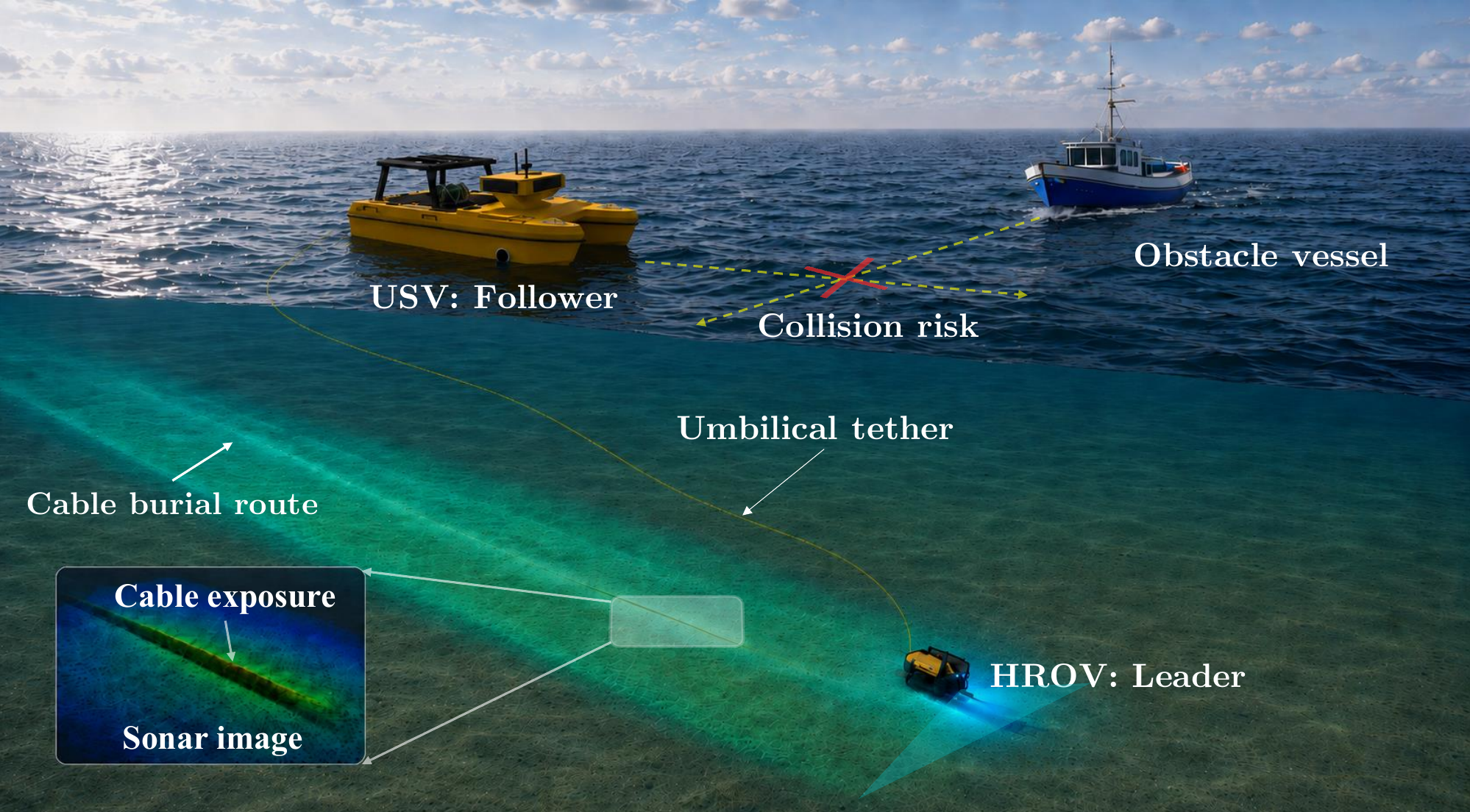}}
	\caption{Illustration of USV collision risk during HROV tracking in subsea cable inspection missions.}
	\label{figure1}
\end{figure}

In recent years, heterogeneous marine robotic systems composed of an unmanned surface vehicle (USV) and a hybrid remotely operated vehicle (HROV) have been increasingly used in subsea cable inspection missions \cite{b1}.
As shown in Fig.~\ref{figure1}, this type of system typically operates in a leader-follower configuration.
The HROV acts as the underwater inspection platform and follows the cable burial route to detect potential hazards, such as cable exposure and free spans \cite{b2}.
Meanwhile, the USV tracks the HROV at the surface and provides power and communication through an umbilical tether \cite{b3}.
However, subsea cable routes are often buried in nearshore waters with frequent vessel traffic. Therefore, the USV is likely to encounter passing surface vessels during HROV tracking, which substantially increases the risk of collision \cite{b4}.
Thus, the USV must avoid passing vessels to ensure overall system safety.

Over the past decade, significant progress has been made in dynamic collision avoidance for USVs.
Most studies have focused on collision avoidance methods that comply with the International Regulations for Preventing Collisions at Sea (COLREGs), demonstrating effectiveness in complex waters \cite{b5}, narrow channels \cite{b6}, and rule-violating scenarios \cite{b7}.
However, these methods are typically designed for point-to-point navigation, with limited consideration of tracking tasks.
Motivated by this demand, several studies have further explored collision avoidance methods for USV tracking scenarios, aiming to ensure safe navigation while maintaining tracking continuity \cite{b8,b9}.
Nevertheless, these methods generally assume that the tracked target is not physically connected to the USV, which limits their applicability to HROV tracking scenarios, where the umbilical tether introduces additional collision risks and motion constraints.
Therefore, dynamic collision avoidance for USVs in HROV tracking scenarios remains a challenging problem and warrants further study.

In practice, USV collision avoidance during HROV tracking presents two main challenges.
First, under gravity and hydrodynamic effects, the flexible tether could form a catenary-like shape that extends laterally beyond the USV hull footprint \cite{b10}. 
In such cases, conventional safety domains that merely maintain a safe distance between the USV hull and obstacle vessels are insufficient \cite{b11}.
As a result, the submerged tether may still scrape against the underwater structures of obstacle vessels, causing tether abrasion or even breakage and thereby compromising system safety.
Although explicit tether modeling can, in principle, characterize such risks in more detail \cite{b12}, 
its computational cost and dependence on parameter settings make it unsuitable for real-time collision avoidance in dynamic environments.
Second, the tether mechanically connects the USV and the HROV, resulting in strong motion coupling.
During USV evasive maneuvers, the USV--HROV distance may increase rapidly, causing the tether to become fully taut \cite{b13}. This condition may lead to HROV deviations from the inspection route or even system damage.
However, existing collision avoidance methods treat the USV as an independently actuated agent, which limits their ability to handle the tether-induced motion coupling and tautness risk in USV-HROV systems \cite{b14}.

To address these issues, this work proposes a tether-aware dynamic collision avoidance method for USV-HROV systems.
First, a tether safety-aware planar domain is introduced to characterize tether-related collision risk without an explicit tether model.
Specifically, ellipsoidal envelopes are constructed for both the tether and an obstacle vessel to represent their three-dimensional occupied regions.
A projection-based dimensional reduction is then introduced to map these regions onto the horizontal plane, thereby constructing planar domains suitable for USV dynamic collision avoidance.
Based on this domain, a tether tautness-aware velocity obstacle (TTA-VO) method is developed.
By incorporating a heuristic tether length management strategy and a tether-release constraint velocity set, the proposed method can generate safe evasive maneuvers while reducing the likelihood of tether tautness.
The TTA-VO method is further integrated with line-of-sight (LOS) guidance to coordinate HROV tracking and dynamic obstacle avoidance.
Gazebo-based simulations demonstrate the effectiveness of the proposed method.
The main contributions of this work are summarized as follows.

1) A tether safety-aware planar domain is proposed to represent, on the horizontal plane, the three-dimensional collision risk between the tether and obstacle vessels without requiring explicit tether modeling.

2) A tether tautness-aware velocity obstacle method is developed by incorporating a heuristic tether length management strategy and a tether-release constraint velocity set, thereby achieving safe avoidance while reducing the likelihood of tether tautness.

3) The proposed components are integrated into a line-of-sight guidance framework to unify target tracking and collision avoidance, thereby maintaining HROV tracking continuity after USV evasive maneuvers.

The remainder of this work is organized as follows.
Section II introduces the tether safety-aware planar domain.
Section III presents the TTA-VO method and the integrated guidance framework.
Section IV demonstrates the simulation results, and Section V concludes this work.

\section{Tether Safety-Aware Planar Domain}
\subsection{Ellipsoidal Envelopes Construction}
\begin{figure}[!t]
\centerline{\includegraphics[width=\columnwidth]{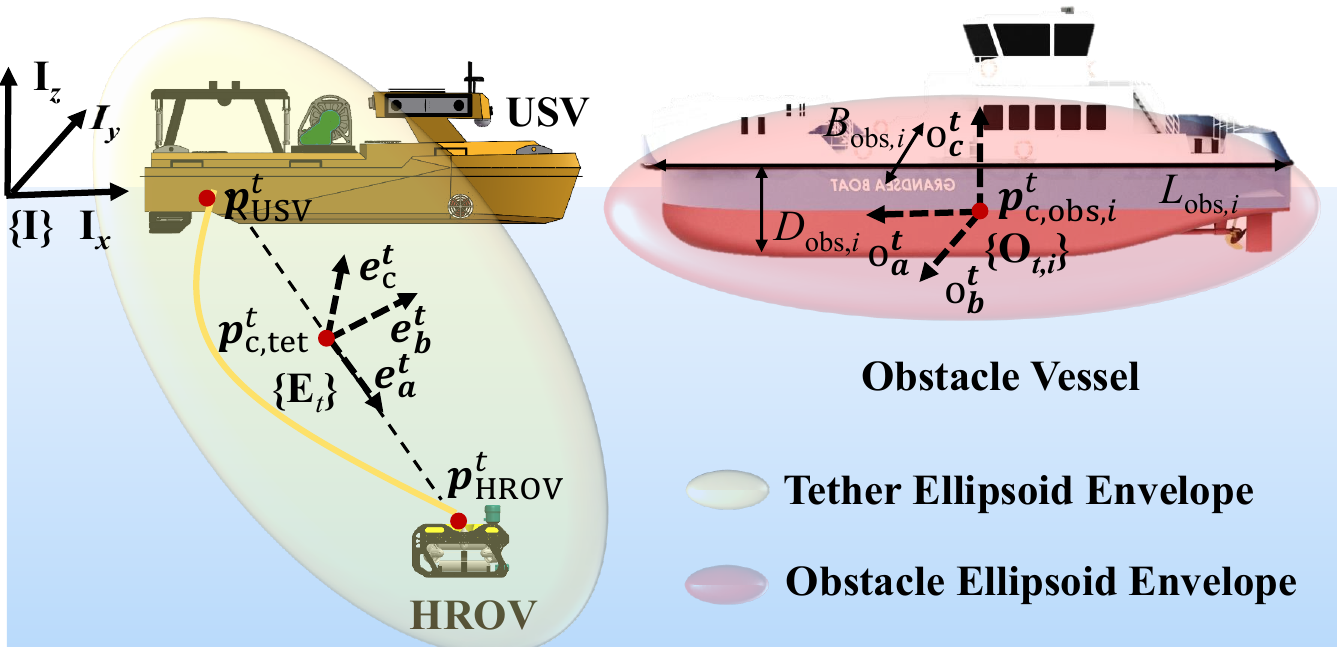}}
\caption{Ellipsoidal envelopes used to bound the tether occupancy region and the obstacle-vessel occupied region.}
\label{figure2}
\end{figure}

\textbf{1) Tether Ellipsoidal Envelope:}
Instead of explicitly reconstructing the tether configuration, a standard ellipsoidal envelope is used to bound the possible tether occupancy region, as illustrated in Fig.~\ref{figure2}.
According to the ellipsoidal geometric definition \cite{b15},
the tether occupancy region can be conservatively modeled by taking 
$\boldsymbol{p}_{\mathrm{USV}}^t$ and $\boldsymbol{p}_{\mathrm{HROV}}^t$ 
as the two foci and using the deployed tether length $L_{\mathrm{tet}}^t$ 
as the prescribed sum of distances, given by
\begin{equation}
\mathcal{E}_{\mathrm{tet}}^t=
\left\{
\boldsymbol{x}\in\mathbb{R}^{3}:
\|\boldsymbol{x}-\boldsymbol{p}_{\mathrm{USV}}^t\|
+
\|\boldsymbol{x}-\boldsymbol{p}_{\mathrm{HROV}}^t\|
\le L_{\mathrm{tet}}^t
\right\}.
\label{eq:tether_ellipse}
\end{equation}

To obtain a compact quadratic form, a local frame aligned with the principal axes of the tether ellipsoid is introduced.
The ellipsoid center is the midpoint of the two foci,
$\boldsymbol{p}_{\mathrm{c,tet}}^{t}=(\boldsymbol{p}_{\mathrm{USV}}^{t}+\boldsymbol{p}_{\mathrm{HROV}}^{t})/2$, and the major-axis direction is
$\boldsymbol{e}_{a}^{t}
=
(\boldsymbol{p}_{\mathrm{HROV}}^{t}-\boldsymbol{p}_{\mathrm{USV}}^{t})/d_{\mathrm{UH}}^{t}$, where $d_{\mathrm{UH}}^t=\|\boldsymbol{p}_{\mathrm{HROV}}^{t}-\boldsymbol{p}_{\mathrm{USV}}^{t}\|$ is the three-dimensional distance between the two tether endpoints.
Two orthonormal directions $\boldsymbol{e}_{b}^{t}$ and $\boldsymbol{e}_{c}^{t}$ are then selected to complete a right-handed basis, and the corresponding rotation matrix is
$\boldsymbol{R}_{E_t}^{I}=[\boldsymbol{e}_{a}^{t},\boldsymbol{e}_{b}^{t},\boldsymbol{e}_{c}^{t}]$.
Equivalently, in the inertial frame, the tether envelope can be written in quadratic form as
\begin{equation}
\mathcal{E}_{\mathrm{tet}}^{t}
=
\left\{
\boldsymbol{x}:
(\boldsymbol{x}-\boldsymbol{p}_{\mathrm{c,tet}}^{t})^{\top}
\boldsymbol{A}_{\mathrm{tet}}^{t}
(\boldsymbol{x}-\boldsymbol{p}_{\mathrm{c,tet}}^{t})
\le 1
\right\},
\label{eq:tether_quad}
\end{equation}
where $\boldsymbol{A}_{\mathrm{tet}}^{t}$ denotes the shape matrix of the tether ellipsoid in the inertial frame $\{I\}$, given by $\boldsymbol{A}_{\mathrm{tet}}^{t}=
\boldsymbol{R}_{E_t}^{I}
\mathrm{diag}(a_{\mathrm{tet}}^{-2},b_{\mathrm{tet}}^{-2},c_{\mathrm{tet}}^{-2})
(\boldsymbol{R}_{E_t}^{I})^{\top}$.
Here, $\boldsymbol{R}_{E_t}^{I}$ is the rotation matrix from the local ellipsoid frame $\{E_t\}$ to the inertial frame. $a_{\mathrm{tet}}$, $b_{\mathrm{tet}}$, and $c_{\mathrm{tet}}$ are the semi-axis lengths of the tether ellipsoid, defined as
\begin{equation}
a_{\mathrm{tet}}=\frac{L_{\mathrm{tet}}^t}{2},\qquad
b_{\mathrm{tet}}=c_{\mathrm{tet}}
=\frac{1}{2}\sqrt{(L_{\mathrm{tet}}^t)^2-(d_{\mathrm{UH}}^t)^2}.
\end{equation}
This formulation assumes $L_{\mathrm{tet}}^t\ge d_{\mathrm{UH}}^t$, which is required for the deployed tether to span the two endpoints.

\textbf{2) Obstacle Ellipsoidal Envelope:}
The $i$th obstacle vessel is approximated by a cuboid with length $L_{\mathrm{obs},i}$, beam $B_{\mathrm{obs},i}$, and draft $D_{\mathrm{obs},i}$.
Its minimum-volume enclosing ellipsoid is used as a conservative outer approximation \cite{b16}, with semi-axes
$a_{\mathrm{obs},i}=\sqrt{3}L_{\mathrm{obs},i}/2$,
$b_{\mathrm{obs},i}=\sqrt{3}B_{\mathrm{obs},i}/2$, and
$c_{\mathrm{obs},i}=\sqrt{3}D_{\mathrm{obs},i}/2$.
Let $\psi_{\mathrm{obs},i}^{t}$ denote the obstacle heading at time $t$.
The rotation matrix from the obstacle body-fixed frame to the inertial frame is defined as
\begin{equation}
\boldsymbol{R}_{O_i}^{I}
=
\begin{bmatrix}
\cos\psi_{\mathrm{obs},i}^{t} & -\sin\psi_{\mathrm{obs},i}^{t} & 0 \\
\sin\psi_{\mathrm{obs},i}^{t} & \cos\psi_{\mathrm{obs},i}^{t} & 0 \\
0 & 0 & 1
\end{bmatrix}.
\end{equation}
The obstacle-envelope shape matrix is then defined as
$\boldsymbol{A}_{\mathrm{obs},i}^{t}
=
\boldsymbol{R}_{O_i}^{I}
\mathrm{diag}(a_{\mathrm{obs},i}^{-2},b_{\mathrm{obs},i}^{-2},c_{\mathrm{obs},i}^{-2})
(\boldsymbol{R}_{O_i}^{I})^{\top}$.
Since $\boldsymbol{p}_{\mathrm{obs},i}^{t}$ denotes the waterline reference position of the obstacle vessel, the ellipsoid center is shifted downward by half the draft, given by
$\boldsymbol{p}_{\mathrm{c,obs},i}^{t}
=
\boldsymbol{p}_{\mathrm{obs},i}^{t}
+[0,0,-D_{\mathrm{obs},i}/2]^{\top}$.
Therefore, the obstacle envelope is expressed as
\begin{equation}
\mathcal{E}_{\mathrm{obs},i}^{t}
=\{\,\boldsymbol{x}\in\mathbb{R}^{3}:(\boldsymbol{x}-\boldsymbol{p}_{\mathrm{c,obs},i}^{t})^{\top}
\boldsymbol{A}_{\mathrm{obs},i}^{t}
(\boldsymbol{x}-\boldsymbol{p}_{\mathrm{c,obs},i}^{t})\le 1\,\}.
\end{equation}

\begin{figure}[!t]
\centerline{\includegraphics[width=\columnwidth]{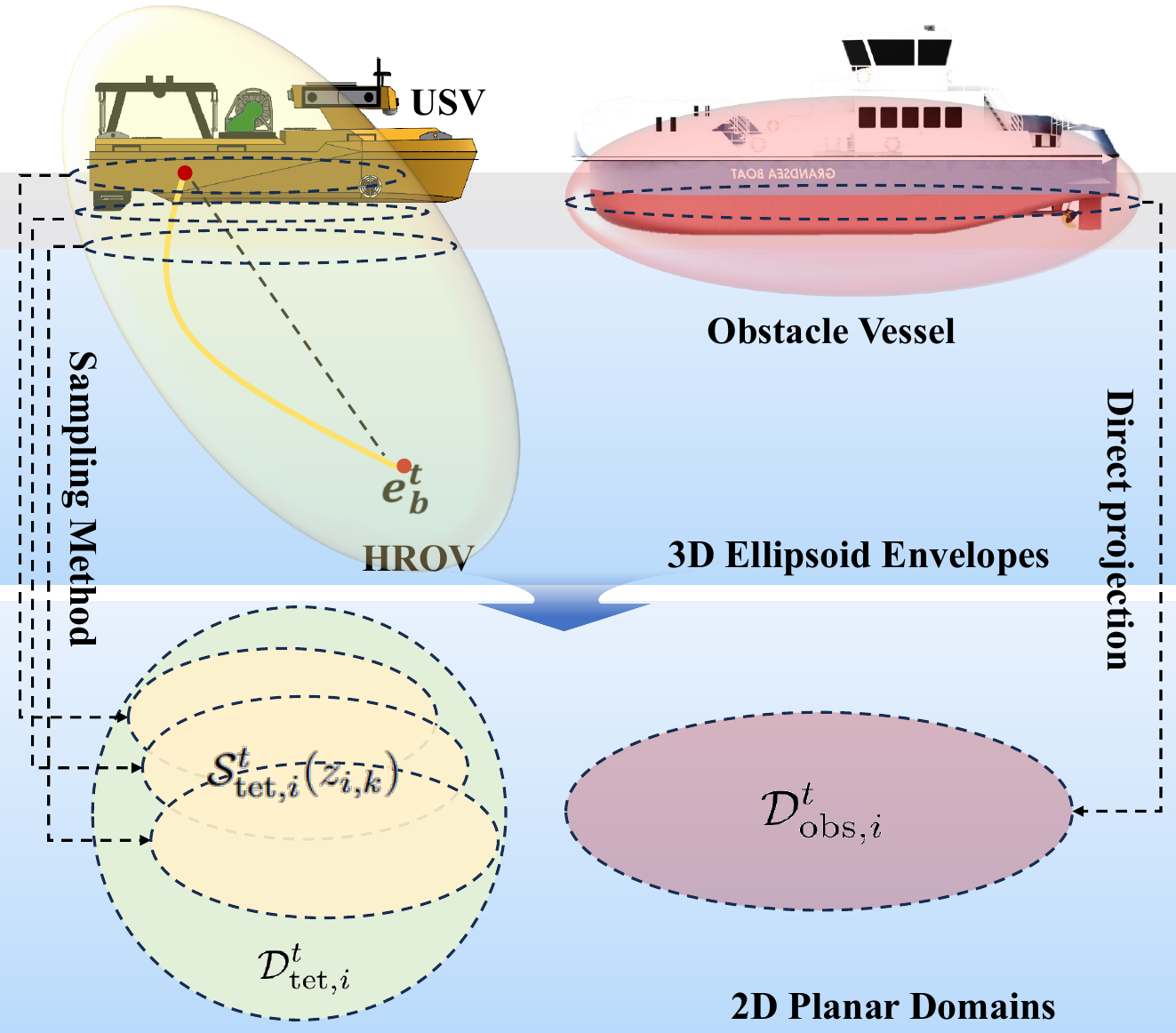}}
\caption{Projection from three-dimensional ellipsoidal envelopes to tether safety-aware planar domains.}
\label{fig:domain}
\end{figure}

\subsection{Planar Domain Construction}
To make the above three-dimensional tether-obstacle collision risk tractable for the subsequent VO-based collision avoidance planning, these envelopes are conservatively mapped into planar domains that preserve the collision-relevant horizontal footprints, as illustrated in Fig.~\ref{fig:domain}.

\textbf{1) Obstacle Planar Domain:}
For the $i$th obstacle vessel, let $\mathcal{S}_{\mathrm{obs},i}^{t}(z)\subset\mathbb{R}^{2}$ denote the horizontal cross-section of its ellipsoidal envelope at depth $z$.
Since the envelope is aligned with the vessel heading and centered at the mid-draft plane, its cross-sections within the draft interval are concentric ellipses, with the largest one located at $z=-D_{\mathrm{obs},i}/2$.
Therefore, the obstacle planar domain is defined by this maximal cross-section, given by
\begin{equation}
\mathcal{D}_{\mathrm{obs},i}^{t}
=
\mathcal{S}_{\mathrm{obs},i}^{t}
\!\left(-\frac{D_{\mathrm{obs},i}}{2}\right).
\label{eq:obs_domain}
\end{equation}

\textbf{2) Tether Planar Domain:}
Unlike the obstacle envelope, the tether ellipsoidal envelope is generally oriented along the USV-HROV connecting direction. 
Consequently, its horizontal cross-sections vary with depth in both position and size, and cannot be replaced by a single maximal cross-section.
Since a tether-obstacle collision can occur only within the submerged draft interval of the obstacle vessel, the collision-relevant depth range is taken as $z\in[-D_{\mathrm{obs},i},0]$.
A uniform depth-sampling strategy is then used to cover the tether cross-sections within this interval. 
Let $N_z\ge2$ denote the number of sampled depth layers, and define
\begin{equation}
z_{i,k}
=
-\frac{k-1}{N_z-1}D_{\mathrm{obs},i},
\qquad
k=1,\ldots,N_z .
\end{equation}
For each sampled depth, $\mathcal{S}_{\mathrm{tet},i}^{t}(z_{i,k})\subset\mathbb{R}^{2}$ denotes the corresponding horizontal cross-section of the tether ellipsoidal envelope.
The union of the boundaries of all sampled cross-sections is defined as
\begin{equation}
\mathcal{Q}_{\mathrm{tet},i}^{t}
=
\bigcup_{k=1}^{N_z}
\partial\!\left(
\mathcal{S}_{\mathrm{tet},i}^{t}(z_{i,k})
\right),
\end{equation}
where $\partial(\cdot)$ denotes the boundary of a planar set.
In implementation, each cross-section boundary is obtained from the quadratic ellipsoid equation at the sampled depth and discretized by uniformly sampling its angular parameter with $N_\theta$ points.

Based on $\mathcal{Q}_{\mathrm{tet},i}^{t}$, a circular planar domain is constructed to represent the tether-related collision region on the horizontal plane.
Its center is fixed at the horizontal position of the USV,
$\boldsymbol{p}_{\mathrm{USV},xy}^{t}=[x_{\mathrm{USV}}^{t},y_{\mathrm{USV}}^{t}]^{\top}$,
and the radius is chosen to cover both the sampled tether cross-sections and the safety extent of the USV hull, given by
\begin{equation}
r_{\mathrm{disc},i}^{t}
=
\max\!\left(
\max_{\boldsymbol{q}\in\mathcal{Q}_{\mathrm{tet},i}^{t}}
\left\|
\boldsymbol{q}-\boldsymbol{p}_{\mathrm{USV},xy}^{t}
\right\|_{2}
+\delta_{\mathrm{samp}},
\;
r_{\mathrm{USV}}^{\mathrm{safe}}
\right),
\end{equation}
where $\delta_{\mathrm{samp}}>0$ is an inflation margin for finite-depth sampling, and $r_{\mathrm{USV}}^{\mathrm{safe}}$ is the safety radius of the USV hull.
The tether planar domain with respect to obstacle vessel $i$ is then defined as
\begin{equation}
\mathcal{D}_{\mathrm{tet},i}^{t}
=
\left\{
\boldsymbol{q}\in\mathbb{R}^{2}:
\|\boldsymbol{q}-\boldsymbol{p}_{\mathrm{USV},xy}^{t}\|_2
\le r_{\mathrm{disc},i}^{t}
\right\},
\label{eq:tether_domain}
\end{equation}
By fixing the domain center at the USV position, the sampled tether occupancy and USV hull safety margin are converted into an enlarged USV-centered safety region, which allows the subsequent VO formulation to retain a standard agent-centered representation with only an online radius update.

\section{Tether Tautness-Aware VO Method}
\begin{algorithm}[!t]
	\caption{TTA-VO collision avoidance process}
	\label{alg:tta_vo}
	\small
	\textbf{Input:} $\boldsymbol{p}_{\mathrm{USV}}^t$, $\boldsymbol{v}_{\mathrm{USV},xy}^t$, $\boldsymbol{p}_{\mathrm{HROV}}^t$, $\boldsymbol{v}_{\mathrm{HROV},xy}^t$, $\boldsymbol{p}_{\mathrm{obs},i}^t$, $\boldsymbol{v}_{\mathrm{obs},i}^t$,
	$\psi_{\mathrm{obs},i}^t$, $\boldsymbol{v}_{\mathrm{nom}}^t$, $\mathcal{V}_{\mathrm{adm}}^t$, $L_{\mathrm{tet}}^0$, $L_{\mathrm{tet}}^{\max}$, $\eta_{\mathrm{rel}}$, $\dot L_{\mathrm{rel}}$, $\dot L_{\mathrm{rec}}$, $\Delta\psi_{\max}$\\
	\textbf{Output:} $\boldsymbol{v}_{\mathrm{safe}}^t$ and $L_{\mathrm{tet}}^t$ at each planning step
	
	\textbf{Initialization:} Set $L_{\mathrm{tet}}^0$ as the nominal deployed tether length.
	
	\For{each planning step $t$}{
		\textbf{Set} $\mathcal{V}_{\mathrm{forb}}^t=\emptyset$.\\
		\For{each obstacle vessel $i\in\mathcal{I}_{\mathrm{risk}}^t$}{
			\textbf{Update} the tether and obstacle ellipsoidal envelopes, and construct the corresponding planar domains $\mathcal{D}_{\mathrm{tet},i}^{t}$ and $\mathcal{D}_{\mathrm{obs},i}^{t}$ using \eqref{eq:obs_domain} and \eqref{eq:tether_domain}.\\
			\textbf{Expand} $\mathcal{D}_{\mathrm{obs},i}^{t}$ outward by the tether-domain radius $r_{\mathrm{disc},i}^{t}$ to obtain $\bar{\mathcal{D}}_{\mathrm{obs},i}^{t}$.\\
			\textbf{Construct} the finite-horizon VO set $\mathcal{V}_{\mathrm{obs},i}^{t}$ using \eqref{eq:vo} and update $\mathcal{V}_{\mathrm{forb}}^t$.
		}
		
		\textbf{Compute} the preliminary safe velocity set
		$
		\mathcal{V}_{\mathrm{pre}}^t
		=
		\mathcal{V}_{\mathrm{adm}}^t\setminus\mathcal{V}_{\mathrm{forb}}^t.
		$
		
		\If{$\boldsymbol{v}_{\mathrm{nom}}^t\notin\mathcal{V}_{\mathrm{forb}}^t$}{
			\textbf{Keep} the nominal tracking velocity:
			$
			\boldsymbol{v}_{\mathrm{safe}}^t
			=
			\boldsymbol{v}_{\mathrm{nom}}^t.
			$ \\
			\If{$L_{\mathrm{tet}}^{t-\Delta t}>L_{\mathrm{tet}}^0$}{
				\textbf{Recover} tether according to \eqref{eq:tether_update}.
			}
			\Else{
				\textbf{Maintain} tether length:
				$
				L_{\mathrm{tet}}^t
				=
				L_{\mathrm{tet}}^{t-\Delta t}.
				$
			}
		}
		\Else{
			\If{$d_{\mathrm{UH}}^t \ge \eta_{\mathrm{rel}}L_{\mathrm{tet}}^{t-\Delta t}$}{
				\textbf{Release} tether according to \eqref{eq:tether_update} and construct $\mathcal{V}_{\mathrm{tet}}^t$ using \eqref{eq:tether_velocity}.\\
				\textbf{Set} $\mathcal{V}_{\mathrm{safe}}^t=\mathcal{V}_{\mathrm{pre}}^t\cap\mathcal{V}_{\mathrm{tet}}^t$.
			}
			\Else{
				\textbf{Maintain} tether length:
				$
				L_{\mathrm{tet}}^t
				=
				L_{\mathrm{tet}}^{t-\Delta t}
				$ \\
				\textbf{Set} $\mathcal{V}_{\mathrm{safe}}^t=\mathcal{V}_{\mathrm{pre}}^t$.
			}
			\textbf{Select} $\boldsymbol{v}_{\mathrm{safe}}^t$ using \eqref{eq:select}.
		}
	}
	
	\textbf{Return:} $\boldsymbol{v}_{\mathrm{safe}}^t$ and $L_{\mathrm{tet}}^t$ at each planning step.
\end{algorithm}
\subsection{Implementation Process}
The implementation process of the TTA-VO method is summarized in Algorithm~\ref{alg:tta_vo}.
At each planning step, the tether safety-aware planar domain is constructed using $L_{\mathrm{tet}}^{t-\Delta t}$ and the current USV-HROV configuration.
For each obstacle vessel in the risk set, the obstacle planar domain $\mathcal{D}_{\mathrm{obs},i}^{t}$ is expanded outward by the radius of the circular tether planar domain $r_{\mathrm{disc},i}^{t}$, defined as
\begin{equation}
\begin{aligned}
\bar{\mathcal{D}}_{\mathrm{obs},i}^{t}
&=
\mathcal{D}_{\mathrm{obs},i}^{t}
\oplus
\mathcal{B}\!\left(r_{\mathrm{disc},i}^{t}\right) \\
&=
\left\{
\boldsymbol{r}\in\mathbb{R}^{2}:
\mathrm{dist}\!\left(\boldsymbol{r},\mathcal{D}_{\mathrm{obs},i}^{t}\right)
\le r_{\mathrm{disc},i}^{t}
\right\},
\end{aligned}
\label{eq:expanded_obs_domain}
\end{equation}
where $\oplus$ denotes the Minkowski sum, and $\mathcal{B}(r_{\mathrm{disc},i}^{t})=\{\boldsymbol{s}\in\mathbb{R}^{2}:\|\boldsymbol{s}\|_2\le r_{\mathrm{disc},i}^{t}\}$ is a disk centered at the origin.
Here, $\boldsymbol{r}$ denotes a point in the obstacle-centered relative plane, and $\mathrm{dist}(\boldsymbol{r},\mathcal{D}_{\mathrm{obs},i}^{t})$ denotes the Euclidean point-to-set distance from $\boldsymbol{r}$ to the obstacle planar domain.
For a prediction horizon $T_h$, the obstacle velocity is assumed to remain constant within the prediction window, and the finite-horizon VO set of obstacle vessel $i$ is defined as
\begin{equation}
	\mathcal{V}_{\mathrm{obs},i}^{t}
	=
	\left\{
	\boldsymbol{v}:
	\exists\tau\in(0,T_h],
	\boldsymbol{r}_i^{t}(\tau,\boldsymbol{v})
	\in
	\bar{\mathcal{D}}_{\mathrm{obs},i}^{t}
	\right\},
	\label{eq:vo}
\end{equation}
where
$\boldsymbol{r}_i^{t}(\tau,\boldsymbol{v})
=
\boldsymbol{p}_{\mathrm{USV},xy}^{t}
-\boldsymbol{p}_{\mathrm{obs},i,xy}^{t}
+\tau(\boldsymbol{v}-\boldsymbol{v}_{\mathrm{obs},i}^{t})$.
The VO sets of nearby obstacle vessels are collected as the forbidden velocity region,
\begin{equation}
	\mathcal{V}_{\mathrm{forb}}^t=
	\bigcup_{i\in\mathcal{I}_{\mathrm{risk}}^t}
	\mathcal{V}_{\mathrm{obs},i}^{t},
\end{equation}
where $\mathcal{I}_{\mathrm{risk}}^t=\{i:d_i^t<d_{\mathrm{risk}}\}$,
$d_i^t=\|\boldsymbol{p}_{\mathrm{obs},i,xy}^{t}
-\boldsymbol{p}_{\mathrm{USV},xy}^{t}\|_2$.

\begin{figure*}[!t]
	\centerline{\includegraphics[width=0.9\textwidth]{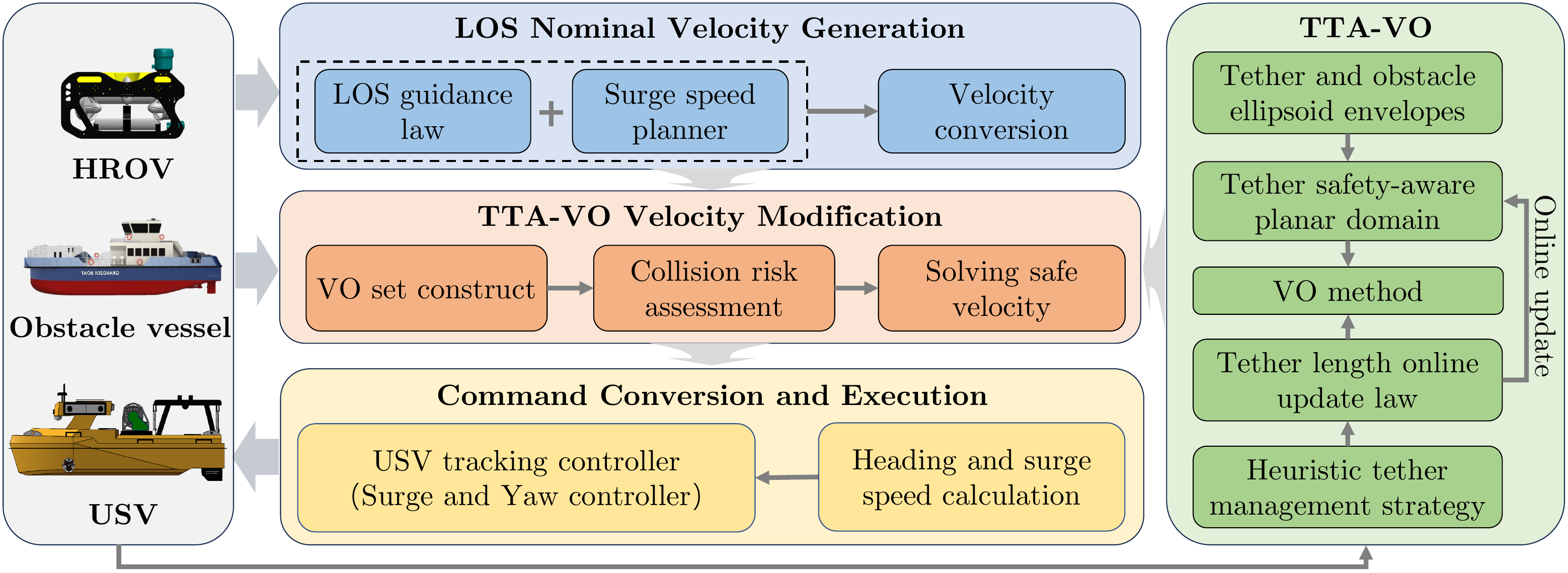}}
	\caption{Integration of LOS tracking guidance and tether-aware dynamic collision avoidance.}
	\label{fig:framework}
\end{figure*}

To ensure executable commands, the admissible velocity set incorporates the surge-speed and heading-increment limits before collision checking:
\begin{equation}
\begin{aligned}
\mathcal{V}_{\mathrm{adm}}^t
=
\{\,\boldsymbol{v}\in\mathbb{R}^{2}:&
\ u_{\min}\le \|\boldsymbol{v}\|_2\le u_{\max},\\
&
|\mathrm{wrap}(\psi_v-\psi_{\mathrm{cmd}}^{t-\Delta t})|
\le \Delta\psi_{\max}
\,\},
\end{aligned}
\end{equation}
where $\psi_v=\mathrm{atan2}(v_y,v_x)$ is the heading angle associated with the candidate velocity, $\mathrm{wrap}(\cdot)$ maps an angle to $(-\pi,\pi]$, $\psi_{\mathrm{cmd}}^{t-\Delta t}$ is the previous heading command, and $\Delta\psi_{\max}$ defines the heading-increment bound.
Accordingly, the preliminary collision-free velocity candidates are obtained as
\begin{equation}
\mathcal{V}_{\mathrm{pre}}^t
=
\mathcal{V}_{\mathrm{adm}}^t\setminus\mathcal{V}_{\mathrm{forb}}^t .
\end{equation}
The nominal velocity is used directly if $\boldsymbol{v}_{\mathrm{nom}}^t\notin\mathcal{V}_{\mathrm{forb}}^t$. Otherwise, an evasive velocity is selected from $\mathcal{V}_{\mathrm{pre}}^t$.

During avoidance, tether release is triggered only when the USV-HROV distance approaches the deployed tether length:
\begin{equation}
\mathcal{R}_{\mathrm{rel}}^t:
\quad
d_{\mathrm{UH}}^t \ge \eta_{\mathrm{rel}}L_{\mathrm{tet}}^{t-\Delta t},\quad
\boldsymbol{v}_{\mathrm{nom}}^t\in\mathcal{V}_{\mathrm{forb}}^t ,
\end{equation}
where $\eta_{\mathrm{rel}}\in(0,1)$ is the release threshold.
When the nominal velocity becomes collision-free, the released tether is recovered:
\begin{equation}
\mathcal{R}_{\mathrm{rec}}^t:
\quad
L_{\mathrm{tet}}^{t-\Delta t}>L_{\mathrm{tet}}^0,\quad
\boldsymbol{v}_{\mathrm{nom}}^t\notin\mathcal{V}_{\mathrm{forb}}^t .
\end{equation}
The deployed tether length is updated as
\begin{equation}
L_{\mathrm{tet}}^t=
\begin{cases}
\min\{L_{\mathrm{tet}}^{\max},L_{\mathrm{tet}}^{t-\Delta t}+\dot L_{\mathrm{rel}}\Delta t\},
& \mathcal{R}_{\mathrm{rel}}^t,\\
\max\{L_{\mathrm{tet}}^0,L_{\mathrm{tet}}^{t-\Delta t}-\dot L_{\mathrm{rec}}\Delta t\},
& \mathcal{R}_{\mathrm{rec}}^t,\\
L_{\mathrm{tet}}^{t-\Delta t},&\text{otherwise},
\end{cases}
\label{eq:tether_update}
\end{equation}
where $L_{\mathrm{tet}}^0$, $L_{\mathrm{tet}}^{\max}$, $\dot L_{\mathrm{rel}}$, $\dot L_{\mathrm{rec}}$, and $\Delta t$ denote the nominal length, maximum length, release rate, recovery rate, and planning period, respectively.
The updated $L_{\mathrm{tet}}^t$ is stored for the next domain update, while $L_{\mathrm{ava}}^t$ accounts for the release available within the current step.

When tether release is triggered, an additional velocity constraint limits the horizontal separation rate according to the available tether margin and one-step release capability.
Let
\begin{equation}
\boldsymbol{e}_{\mathrm{UH}}^{t}
=
\frac{
\boldsymbol{p}_{\mathrm{USV},xy}^{t}-\boldsymbol{p}_{\mathrm{HROV},xy}^{t}}
{\|\boldsymbol{p}_{\mathrm{USV},xy}^{t}-\boldsymbol{p}_{\mathrm{HROV},xy}^{t}\|_2}
\end{equation}
be the horizontal separation direction from the HROV to the USV, with the previous valid direction used when $d_{\mathrm{UH},xy}^t=0$. The tether-constrained velocity set is
\begin{equation}
\mathcal{V}_{\mathrm{tet}}^t=
\left\{
\boldsymbol{v}\in\mathcal{V}_{\mathrm{adm}}^t:
(\boldsymbol{e}_{\mathrm{UH}}^{t})^{\top}
(\boldsymbol{v}-\boldsymbol{v}_{\mathrm{HROV},xy}^{t})
\le c_{\mathrm{xy}}^{t}
\right\},
\label{eq:tether_velocity}
\end{equation}
where
\begin{align}
c_{\mathrm{xy}}^{t}
&=
\frac{
\sqrt{(L_{\mathrm{ava}}^t)^2-(h_{\mathrm{HROV}}^t)^2}
-d_{\mathrm{UH},xy}^{t}}
{\Delta t}, \\
L_{\mathrm{ava}}^t
&=
\min\{L_{\mathrm{tet}}^{\max},
L_{\mathrm{tet}}^{t-\Delta t}+\dot L_{\mathrm{rel}}\Delta t\}.
\end{align}
Here, $\boldsymbol{v}_{\mathrm{HROV},xy}^{t}$ is the HROV horizontal velocity, $d_{\mathrm{UH},xy}^{t}=\|\boldsymbol{p}_{\mathrm{USV},xy}^{t}-\boldsymbol{p}_{\mathrm{HROV},xy}^{t}\|_2$, $h_{\mathrm{HROV}}^t=|z_{\mathrm{USV}}^t-z_{\mathrm{HROV}}^t|$, $L_{\mathrm{ava}}^t$ is the tether length available within the current step, and $c_{\mathrm{xy}}^{t}$ is the allowable horizontal separation rate.
Thus, the final evasive candidate set is
\begin{equation}
\mathcal{V}_{\mathrm{safe}}^t
=
\begin{cases}
\mathcal{V}_{\mathrm{pre}}^t\cap \mathcal{V}_{\mathrm{tet}}^t,& \mathcal{R}_{\mathrm{rel}}^t,\\
\mathcal{V}_{\mathrm{pre}}^t,& \text{otherwise},
\end{cases}
\end{equation}
The safe velocity command is selected as
\begin{equation}
\boldsymbol{v}_{\mathrm{safe}}^t
=
\begin{cases}
\boldsymbol{v}_{\mathrm{nom}}^t,&
\boldsymbol{v}_{\mathrm{nom}}^t\notin\mathcal{V}_{\mathrm{forb}}^t,\\[1mm]
\displaystyle
\arg\min_{\boldsymbol{v}\in\mathcal{V}_{\mathrm{safe}}^t}
J(\boldsymbol{v}),&
\boldsymbol{v}_{\mathrm{nom}}^t\in\mathcal{V}_{\mathrm{forb}}^t,\ \mathcal{V}_{\mathrm{safe}}^t\ne\emptyset,\\[1mm]
\boldsymbol{v}_{\mathrm{emg}}^t,&
\boldsymbol{v}_{\mathrm{nom}}^t\in\mathcal{V}_{\mathrm{forb}}^t,\ \mathcal{V}_{\mathrm{safe}}^t=\emptyset,
\end{cases}
\label{eq:select}
\end{equation}
with
$J(\boldsymbol{v})=
w_1\|\boldsymbol{v}-\boldsymbol{v}_{\mathrm{nom}}^t\|_2^2+
w_2\|\boldsymbol{v}-\boldsymbol{v}_{\mathrm{USV},xy}^t\|_2^2$.
Here, $w_1,w_2>0$ penalize tracking deviation and abrupt velocity changes, respectively.
$\boldsymbol{v}_{\mathrm{emg}}^t$ denotes an emergency fallback command used when no admissible safe velocity is available. In this work, it is defined as a conservative braking command with zero surge speed.

\subsection{Integration With LOS Guidance}
As shown in Fig.~\ref{fig:framework}, the proposed TTA-VO planner is embedded into the LOS tracking framework as a supervisory velocity-modification layer between the nominal guidance law and the low-level USV controller.
The LOS guidance law first provides a nominal tracking velocity for following the HROV, and the TTA-VO module then checks whether this velocity satisfies the constraints imposed by the tether safety-aware planar domain, the finite-horizon VO formulation, and the tether-tautness condition.
If the nominal velocity is safe, it is directly passed to the command-conversion stage.
Otherwise, the TTA-VO module replaces it with the closest admissible safe velocity according to \eqref{eq:select}.
In this way, the LOS tracking behavior is preserved during normal operation, while collision avoidance and tether-length management are activated only when required.

\textbf{1) LOS Nominal Velocity Generation:}
At each planning step, the LOS guidance law generates the nominal heading from the USV to the HROV:
\begin{equation}
\psi_{\mathrm{nom}}^t=\mathrm{atan2}(y_{\mathrm{HROV}}^t-y_{\mathrm{USV}}^t,
x_{\mathrm{HROV}}^t-x_{\mathrm{USV}}^t),
\end{equation}
and the nominal surge speed is adjusted by the horizontal USV-HROV distance error.
Thus, the nominal tracking velocity is
\begin{equation}
\boldsymbol{v}_{\mathrm{nom}}^t
=
\mathrm{sat}_{[u_{\min},u_{\max}]}
(u_0+k_d(d_{\mathrm{UH},xy}^t-d_{\mathrm{ref}}))
\begin{bmatrix}
\cos\psi_{\mathrm{nom}}^t\\
\sin\psi_{\mathrm{nom}}^t
\end{bmatrix},
\end{equation}
where $d_{\mathrm{UH},xy}^t$ is the horizontal USV-HROV distance defined above, $d_{\mathrm{ref}}$ is the desired tracking distance, $u_0$ is the nominal surge speed, $k_d$ is the distance-error feedback gain, $u_{\min}$ and $u_{\max}$ are the lower and upper surge-speed limits, and $\mathrm{sat}_{[u_{\min},u_{\max}]}(\cdot)$ clips its argument to $[u_{\min},u_{\max}]$.

\textbf{2) TTA-VO Velocity Modification:}
After $\boldsymbol{v}_{\mathrm{nom}}^t$ is generated, the TTA-VO module updates the tether safety-aware planar domain and constructs the forbidden velocity region from the obstacle planar domains.
The nominal velocity is then used as the preferred command in the velocity-selection process.
When $\boldsymbol{v}_{\mathrm{nom}}^t\notin\mathcal{V}_{\mathrm{forb}}^t$, the LOS command is retained and the released tether, if any, is recovered toward the nominal length.
When $\boldsymbol{v}_{\mathrm{nom}}^t\in\mathcal{V}_{\mathrm{forb}}^t$, the admissible velocity set is filtered by the VO constraints and, when necessary, by the tether-release velocity constraint.
The resulting command $\boldsymbol{v}_{\mathrm{safe}}^t$ is selected from the feasible candidates by minimizing its deviation from both the LOS nominal velocity and the current USV velocity.
Therefore, the avoidance planner modifies the LOS command only to the extent needed to maintain obstacle clearance and reduce tether-tautness risk.

\textbf{3) Command Conversion and Execution:}
The selected planar velocity is converted into the heading and surge-speed commands according to its direction and magnitude.
Because the heading-increment limit has already been embedded in $\mathcal{V}_{\mathrm{adm}}^t$, the generated command remains consistent with the executable heading constraint considered during VO planning.
The resulting commands are then sent to the low-level USV controller for execution.

\begin{figure*}[!t]
	\centerline{\includegraphics[width=0.92\textwidth]{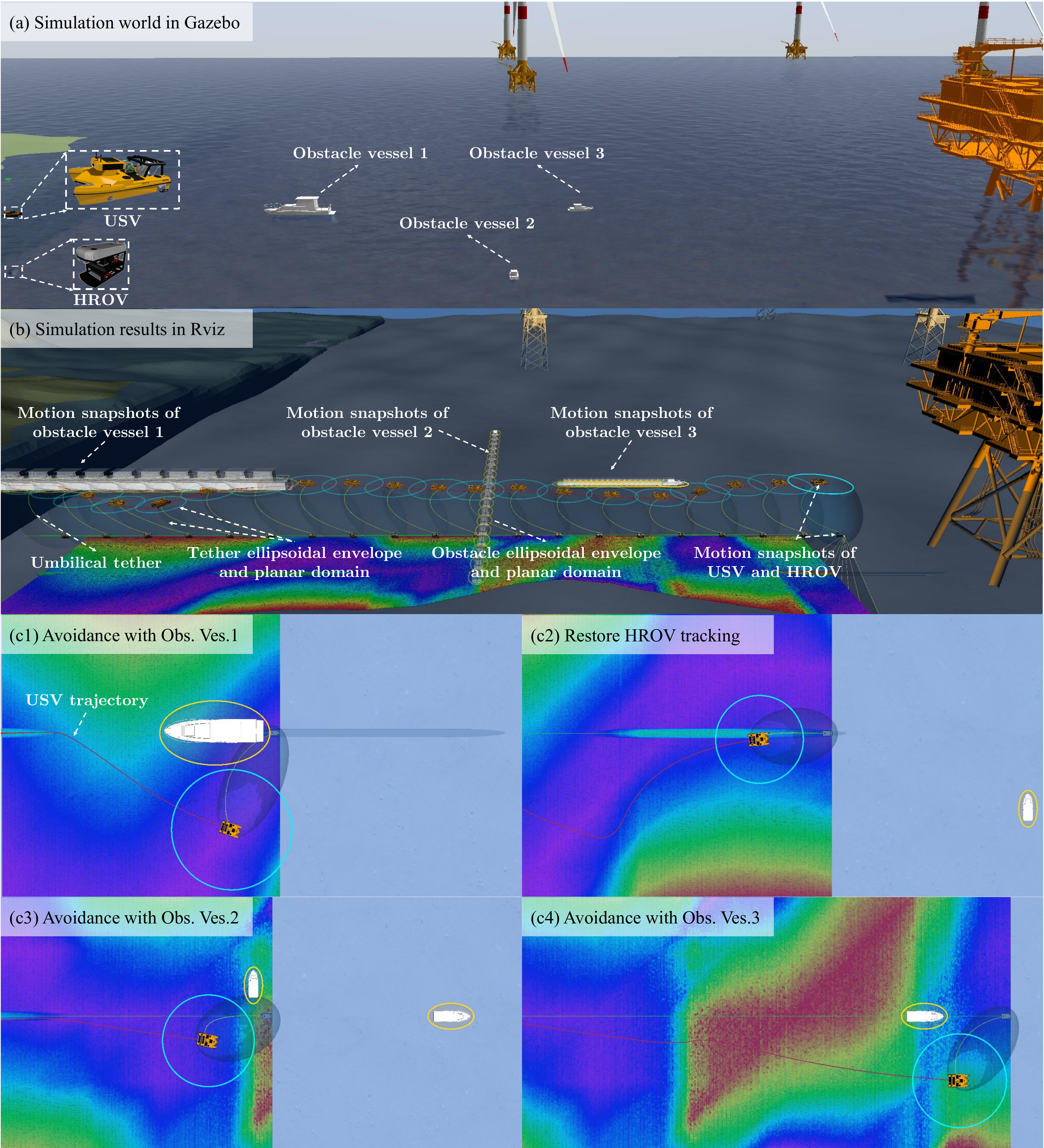}}
	\caption{Gazebo snapshots of the multi-vessel collision avoidance process.}
	\label{figure5}
\end{figure*}
\begin{figure}[!t]
	\centerline{\includegraphics[width=1.0\columnwidth]{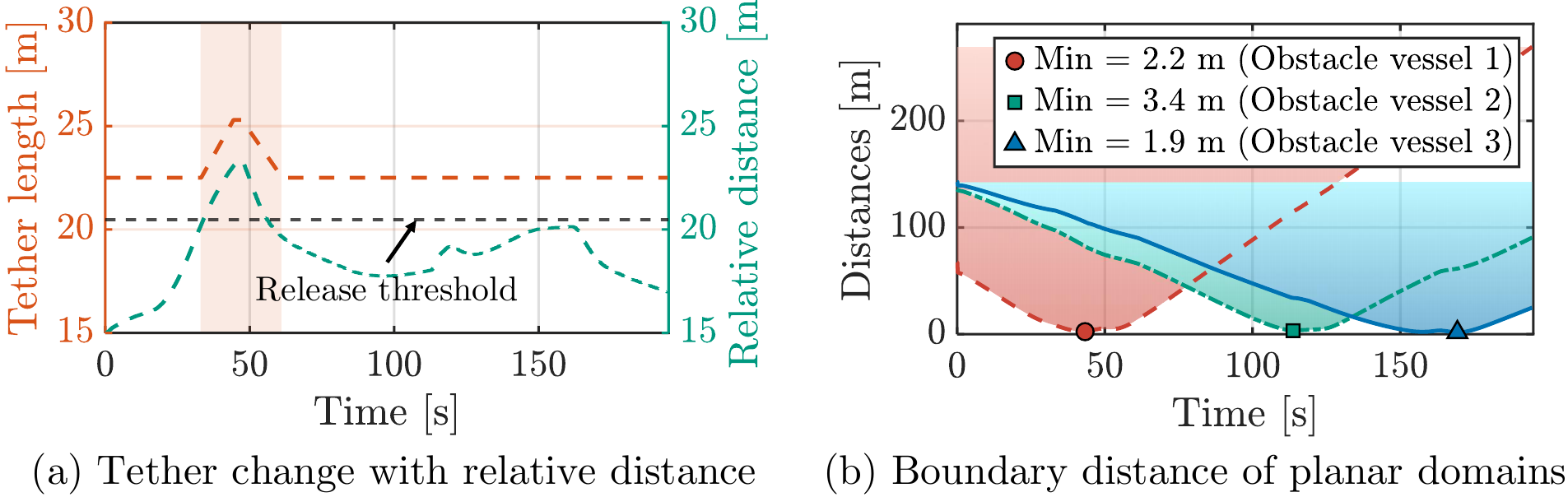}}
	\caption{Simulation results in three typical collision avoidance scenarios.}
	\label{figure6}
\end{figure}
\begin{figure*}[!t]
	\centerline{\includegraphics[width=0.9\textwidth]{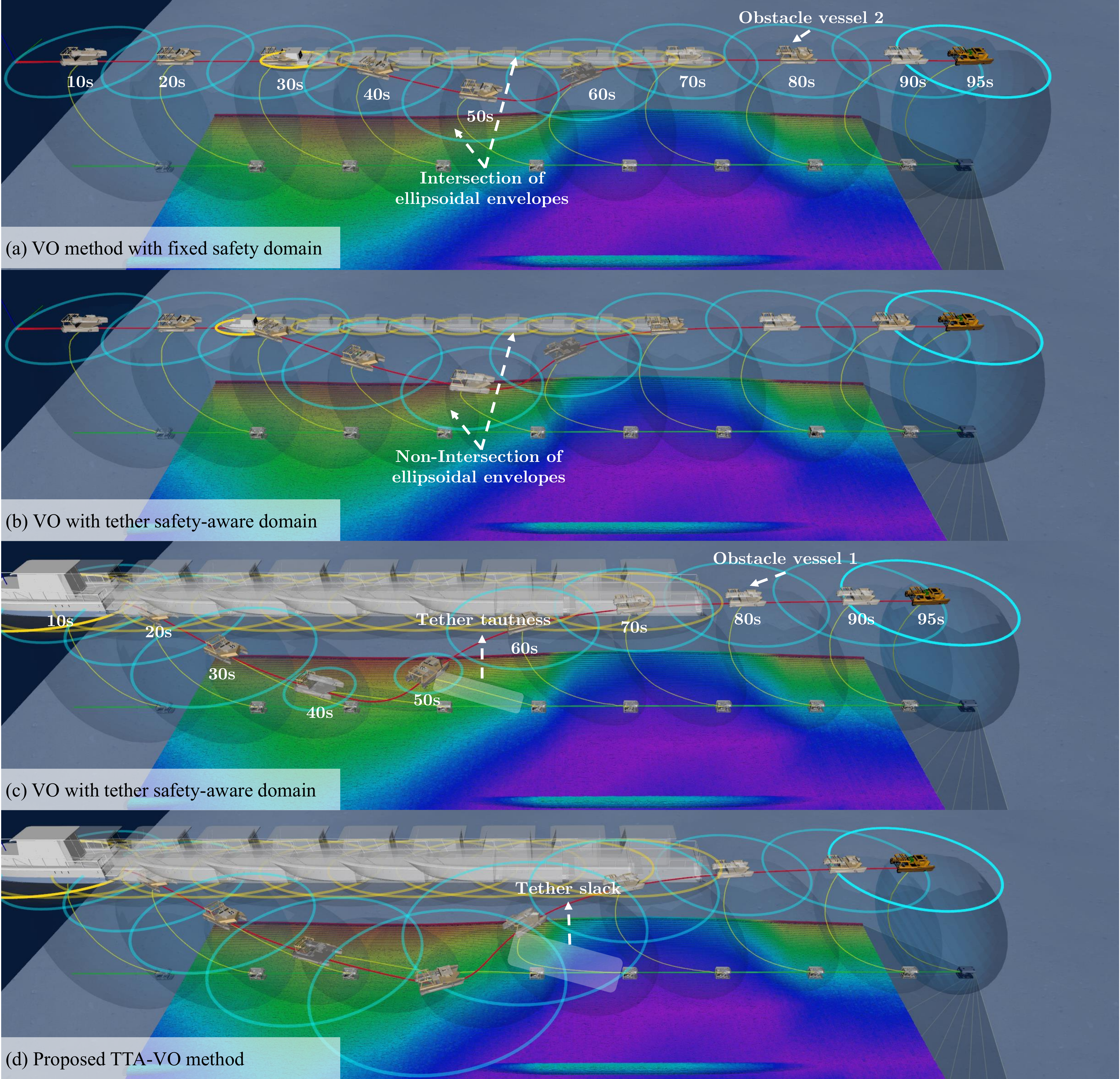}}
	\caption{Ablation simulation results. Evaluation metrics: $d_{\mathrm{bd}}^{\min}$ is the minimum boundary distance between the tether planar domain and the obstacle planar domain;
		$m_{\mathrm{tet}}^{\min}=\min_t(L_{\mathrm{tet}}^t-d_{\mathrm{UH}}^t)$ is the minimum tether-length margin.}
	\label{figure7}
\end{figure*}

\section{Simulation Results}

\subsection{Parameter Settings}
The initial positions were $\boldsymbol{p}_{\mathrm{USV}}^0=[0,0,0]^{\top}$ and $\boldsymbol{p}_{\mathrm{HROV}}^0=[0,0,-15]^{\top}$, and the HROV moved along the inspection route at a constant depth of 15~m with a forward speed of 1.2~m/s.
When $d_{\mathrm{UH},xy}^t=0$, the nominal LOS heading and the initial heading command were initialized along the inspection-route direction.
The obstacle vessels were assumed to move with constant speeds and headings, as listed in Table~\ref{tab:obs_params}.
The nominal tether length was $L_{\mathrm{tet}}^0=22.5$ m, corresponding to 1.5 times the HROV operating depth as a practical winch setting \cite{b17}, and the maximum allowable length was $L_{\mathrm{tet}}^{\max}=100$ m. 
The release threshold was $\eta_{\mathrm{rel}}=0.9$, with release and recovery rates $\dot L_{\mathrm{rel}}=0.35$ m/s and $\dot L_{\mathrm{rec}}=0.2$ m/s, respectively. 
The prediction horizon was $T_h=8.5$ s, the obstacle screening distance was $d_{\mathrm{risk}}=30$ m, and the velocity search used a maximum USV speed of 1.3 m/s. 
To impose a starboard-side avoidance preference consistent with maritime collision-avoidance practice, the candidate headings were sampled only within the right-turn admissible range, with a resolution of $0.5^{\circ}$ and 150 heading samples.
For planar-domain construction, $N_z=10$, $N_\theta=360$, $\delta_{\mathrm{samp}}=1.0$ m, and $r_{\mathrm{USV}}^{\mathrm{safe}}=5$ m were used. 
In the LOS tracking module, $u_0=1.2$ m/s, $k_d=0.25$, $u_{\min}=0.1$ m/s, $u_{\max}=1.3$ m/s, and the desired horizontal tracking distance was $d_{\mathrm{ref}}=1$ m. 
The velocity-selection weights were $w_1=0.8$ and $w_2=0.4$, the planning period was $\Delta t=0.1$ s, and the heading-increment bound was $\Delta\psi_{\max}=0.1$ rad. 

\begin{table}[!t]
\caption{Obstacle Vessel Parameters}
\label{tab:obs_params}
\centering
\scriptsize
\renewcommand{\arraystretch}{1.08}
\setlength{\tabcolsep}{2.5pt}
\begin{tabular}{ccccccc}
\toprule
Num. & $L_{\mathrm{obs},i}$ & $B_{\mathrm{obs},i}$ & $D_{\mathrm{obs},i}$ & $v_{\mathrm{obs},i}$ & $\psi_{\mathrm{obs},i}^0$ & $\boldsymbol{p}_{\mathrm{obs},i}^0$\\
\midrule
1 & 12 m & 7 m & 2.0 m & 0.75 m/s & $180^{\circ}$ & $[75,0,0]$\\
2 & 4.0 m & 2.0 m & 1.0 m & 0.55 m/s & $90^{\circ}$ & $[130,-60,0]$\\
3 & 4.5 m & 2.5 m & 1.5 m & 0.15 m/s & $0^{\circ}$ & $[150,0,0]$\\
\bottomrule
\end{tabular}
\end{table}

\subsection{Multi-Vessel Continuous Avoidance}
Fig.~\ref{figure5}(a) shows the Gazebo simulation world with three dynamic obstacle vessels of different sizes and speeds, while Fig.~\ref{figure5}(b) presents the complete RViz visualization with motion snapshots at 10~s intervals.
Representative avoidance stages are shown in Fig.~\ref{figure5}(c1)--(c4).
During these stages, the circular tether planar domain remains disjoint from the elliptical obstacle planar domain, indicating that the tether safety-aware planar domain provides a geometric basis for enforcing tether safety.
Meanwhile, the USV gradually resumes HROV tracking after the evasive maneuver, as shown in Fig.~\ref{figure5}(c2), demonstrating the coordination between HROV tracking and collision avoidance.
The key quantitative results are shown in Fig.~\ref{figure6}.
Fig.~\ref{figure6}(a) illustrates the effect of the heuristic tether length management strategy.
During the avoidance maneuver for obstacle vessel 1, the evasive detour increases the USV--HROV relative distance beyond the release threshold, triggering tether release.
The deployed tether length increases from the nominal 22.5~m to approximately 25~m, providing additional maneuvering margin to reduce the risk of tether tautness.
After the collision risk is removed, the tether is gradually recovered to its nominal length, indicating that the strategy enlarges the avoidance margin only when needed.
Fig.~\ref{figure6}(b) further shows that the minimum boundary distances between the tether and obstacle planar domains remain positive, with minimum values of 2.2~m, 3.4~m, and 1.9~m for obstacle vessels 1, 2, and 3, respectively.
This confirms that no overlap occurs during the avoidance maneuvers and that tether-related collision risks are avoided.

\subsection{Comparative Analysis}
Ablation studies were conducted to assess the contributions of the key modules. First, the basic VO method with a fixed domain was compared with the VO method incorporating the tether safety-aware planar domain (4~m). 
As shown in Fig.~\ref{figure7}(a), although the basic VO method prevents the USV hull from colliding with the obstacle vessel, the tether ellipsoidal envelope still intersects the obstacle envelope ($d_{\mathrm{bd}}^{\min}=-2.73~\mathrm{m}$), indicating a potential submerged-tether collision risk. 
After introducing the tether safety-aware planar domain, the two envelopes become disjoint ($d_{\mathrm{bd}}^{\min}=3.48~\mathrm{m}$), as shown in Fig.~\ref{figure7}(b).
Furthermore, the planar-domain VO method was compared with the proposed TTA-VO method to evaluate the heuristic tether length management strategy.
As shown in Fig.~\ref{figure7}(c) and (d), the planar-domain VO method avoids tether-obstacle overlap, but the USV--HROV distance may exceed the deployed tether length in the absence of tether release ($m_{\mathrm{tet}}^{\min}=-3.27~\mathrm{m}$), increasing the risk of tether tautness.
In contrast, the proposed TTA-VO method adaptively releases the tether, enlarges the maneuvering margin, and maintains a positive tether-length margin ($m_{\mathrm{tet}}^{\min}=2.55~\mathrm{m}$), thereby preventing the tether from becoming taut during obstacle avoidance.

\section{Conclusion}
This work proposed a tether-aware dynamic collision avoidance method for a USV tracking an HROV during subsea cable inspection. 
To address the risk of submerged tether contact with passing vessels, a tether safety-aware planar domain was constructed by combining ellipsoidal envelope modeling and projection-based dimensional reduction, allowing tether-obstacle collision risk to be represented in a form suitable for online VO planning. 
To reduce the likelihood of tether tautness during evasive maneuvers, a TTA-VO method was developed by incorporating online tether-length management and a tether-release velocity constraint into the velocity selection process. 
The resulting planner was further embedded into an LOS guidance framework, so that nominal HROV tracking is retained when safe and modified only when collision avoidance or tether constraints require it. 
Gazebo simulations demonstrated that the proposed framework avoids dynamic obstacle vessels, maintains positive tether-domain separation, and preserves a positive tether-length margin compared with baseline VO variants. 
Future work will include controlled water-tank experiments, field validation on full-scale USV-HROV platforms, and extension to multi-vessel interaction scenarios with reciprocal behavior.

\end{document}